\documentclass[11pt]{article}

\usepackage[final]{acl}

\usepackage{times}
\usepackage{latexsym}
 \usepackage{booktabs}
 \usepackage{enumitem}

\usepackage[T1]{fontenc}

\usepackage[utf8]{inputenc}

\usepackage{microtype}

\usepackage{inconsolata}

\usepackage{graphicx}
\usepackage{subcaption}

\usepackage[most]{tcolorbox}
\usepackage{listings}
\usepackage{xcolor}
\usepackage{caption}
\usepackage{amssymb}
\usepackage{multirow}
\usepackage[table,dvipsnames]{xcolor}

\usepackage{hyperref}
\usepackage[nameinlink]{cleveref}
\Crefname{figure}{Fig.}{Figs.}
\Crefname{table}{Tab.}{Tabs.}
\usepackage{pifont}
\newcommand{\cmark}{\textcolor{green!60!black}{\ding{51}}}
\newcommand{\xmark}{\textcolor{red}{\ding{55}}}
\definecolor{sectiongray}{RGB}{248,248,248}
\definecolor{oursblue}{RGB}{242,248,255}

\lstdefinestyle{promptstyle}{
    basicstyle=\ttfamily\small,
    columns=fullflexible,
    keepspaces=true,
    breaklines=true,
    breakatwhitespace=false,
    breakautoindent=false,
    breakindent=0pt,
    postbreak={},
    showstringspaces=false,
    frame=none
}

\usepackage[most]{tcolorbox}

\newtcolorbox{promptnewbox}[1]{
    colback=white,
    colframe=black!70,
    coltitle=white,
    colbacktitle=black!70,
    title={#1},
    fonttitle=\bfseries\small,
    fontupper=\small,
    boxrule=0.8pt,
    arc=3pt,
    left=5pt,
    right=5pt,
    top=4pt,
    bottom=4pt,
    before skip=2pt,
    after skip=2pt
}

\title{\method: Manga Active Narrative Grounding Optimization}

\author{
Hao Qiu\textsuperscript{1}\thanks{Equal Contribution.},
Junyan Wang\textsuperscript{2*}\thanks{Corresponding Authors.},
Zheyuan Liu\textsuperscript{2},
Lei Fan\textsuperscript{3},
Hong Jia\textsuperscript{4},
Lianbo Guo\textsuperscript{5},
Zhulin Tao\textsuperscript{1$\dagger$}
\\
\textsuperscript{1}Communication University of China
\\
\textsuperscript{2}Australian Institute for Machine Learning, Adelaide University
\\
\textsuperscript{3}University of New South Wales
\quad
\textsuperscript{4}University of Auckland
\\
\textsuperscript{5}Huazhong University of Science and Technology
\\
\texttt{
qiuhao@cuc.edu.cn
}
\texttt{junyan.wang\&zheyuan.liu@adelaide.edu.au}
\\
\texttt{
lei.fan1@unsw.edu.au
}
\texttt{
hong.jia@auckland.ac.nz
}
\texttt{lbguo@hust.edu.cn}
\\
\texttt{taozl@cuc.edu.cn}
}

\newcommand{\mangavqa}{\text{manga VQA}}
\newcommand{\method}{\textsc{ManGo}}
\newcommand{\activesketch}{ANS}

\begin{document}
\maketitle
\begin{abstract}
Manga visual question answering requires models to answer questions over panel-based visual narratives, where relevant evidence is distributed across ordered panels, embedded text, recurring characters, and implicit event transitions. 
This structure makes passive page encoding insufficient, as the model must identify which panels to inspect, what clues to retain, and when the accumulated evidence is sufficient for answering. 
We propose \method{} (\textbf{M}anga \textbf{A}ctive \textbf{N}arrative \textbf{G}rounding \textbf{O}ptimization), an unsupervised framework for active manga visual question answering.
\method{} introduces Active Narrative Sketching (\activesketch{}), which iteratively selects panels, extracts concise grounded clues, and decides when to stop, forming a compact question-directed evidence sketch before answer generation.
To optimize this behavior without human-annotated answers or rationale paths, \method{} samples multiple \activesketch{} rollouts and applies group-relative training with two rewards: answer preference from listwise self-ranking and path consistency from stable ordered panel trajectories.
The combined reward is optimized with group-relative policy training, encouraging the model to improve both final answers and the panel-level evidence paths that support them.
Experiments on standard manga understanding benchmarks show that \method{} achieves state-of-the-art performance across different settings.
\end{abstract}

\section{Introduction}

\begin{figure}[t]
    \centering
    \includegraphics[width=0.95\columnwidth]{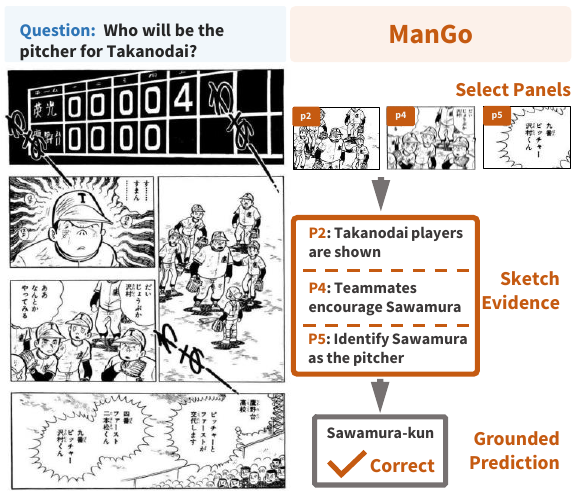}
    \caption{
An illustration of \method{}.
Rather than answering directly from the full manga page, \method{} actively selects question-relevant panels, writes compact panel-grounded clues, and composes them into a narrative evidence sketch. \copyright Mikio Yoshimori
}
    \label{fig:intro}
\end{figure}

We study \mangavqa{}, the task of answering natural-language questions about manga pages. 
Unlike conventional visual question answering over natural images, where the relevant evidence often lies within a single scene, manga pages present a structured visual narrative. 
Information is distributed across ordered panels, embedded text, character appearances, gestures, and implicit transitions between events~\citep{baek2026mangavqa}. 
A correct answer may therefore depend on identifying relevant panels, linking dialogue to recurring characters, and reconstructing temporal or causal relations expressed through the page layout and narrative flow. 
This makes \mangavqa{} less a problem of recognizing isolated visual cues than of understanding distributed, question-dependent evidence in a compact multimodal story.

This form of evidence distribution makes active evidence-seeking a natural fit for \mangavqa{}. 
Active reasoning differs from passive problem solving in that the model must identify and acquire task-relevant evidence before solving the task~\citep{zhoupassive}. 
For manga pages, this means deciding which panel to inspect, what clue to retain, and when the accumulated evidence is sufficient for answering. 
This framing avoids passively encoding the whole page once, and instead matches how \mangavqa{} requires question-relevant evidence to be progressively discovered and connected.

Motivated by this view, we propose \method{} (Manga Active Narrative Grounding Optimization), an unsupervised framework for active manga VQA. 
\method{} consists of two coupled components: Active Narrative Sketching (\activesketch{}) for panel-level evidence construction, and group-relative training for unsupervised policy optimization.
As shown in~\Cref{fig:intro}, given a manga page, panel boxes, and a question, \activesketch{} constructs a structured evidence sketch before answer generation. 
At each step, the model performs active panel perception by selecting a panel under the current question and sketch state, extracting a concise clue grounded in that panel, and deciding whether to continue or answer. 
The resulting narrative evidence sketch is ordered by the evolving evidence need rather than by page order alone, and may revisit a panel when it provides a different useful clue. 
In this way, \activesketch{} turns \mangavqa{} into an iterative process of question-guided panel inspection and evidence-state update.

Prompting alone is unreliable for learning such active sketching behavior.
However, direct supervision is difficult to scale because the model requires process-level supervision, not only final-answer labels. 
Such supervision must specify the sequence of evidence-seeking decisions along the reasoning path: which panel to inspect, which clue to preserve, how to avoid irrelevant details, and when the sketch is sufficient for answering.
We therefore train \method{} with unsupervised reward modeling and group-relative policy optimization, using questions and panel annotations without relying on human-annotated answers or rationale paths.

To obtain training signals without gold-answer supervision, \method{} derives rewards from relations among sampled rollouts. 
For each question, the policy samples multiple \activesketch{} rollouts, each containing an evidence sketch and a final answer. 
Then, we construct two complementary rewards. 
First, answer preference reward estimates answer quality by merging duplicate final answers into candidate clusters and asking the policy model to rank them under the same manga page and question, producing a dense answer-level signal beyond frequency or surface-form agreement. 
Second, path consistency reward evaluates whether a rollout's ordered panel trajectory is supported by stable evidence paths from high-preference rollouts. 
The two rewards are combined and optimized with group-relative policy optimization, encouraging the policy to improve both the final answer and the panel-level evidence path that supports it.

We evaluate \method{} on standard manga understanding benchmarks and show that it achieves state-of-the-art performance across open-ended and multiple-choice settings. 
Overall, our work makes three contributions: 
(1) we propose \method{}, an unsupervised framework for active manga visual question answering; 
(2) we introduce \activesketch{}, which combines active panel perception with narrative evidence sketching to construct compact, question-guided panel evidence paths; and 
(3) we develop an unsupervised reward modeling scheme with answer preference and path consistency rewards, enabling group-relative training without human-annotated answers or rationale paths.

\begin{figure*}[t]
  \includegraphics[width=\linewidth]{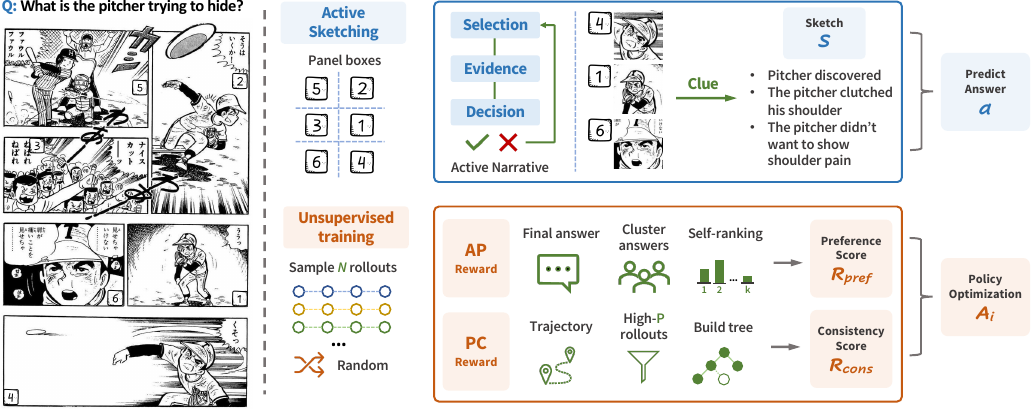}
  \caption{
Overview of \method{}.
The upper part shows Active Narrative Sketching, where the model performs panel selection, evidence sketching, and stopping decisions to build a compact sketch $S$ before answering.
The lower part shows unsupervised training: sampled rollouts are scored by answer preference and path consistency rewards, which are combined for group-relative policy optimization. \copyright Mikio Yoshimori
}
  \label{fig:overview}
\end{figure*}

\section{Background and Preliminaries}
\label{sec:background}

\paragraph{Active Perception.}
Active perception studies how an agent actively acquires task-relevant information instead of passively encoding a fixed observation.
Classical active vision formulates perception as a closed-loop process, where the agent decides what, where, how, or when to sense according to its current goal and observations~\cite{aloimonos1988active}.
Recent multimodal reasoning methods instantiate this idea in MLLMs by allowing the model to select regions, crop images, or zoom into details before answering~\cite{sarch2025grounded, zheng2025deepeyes, shen2025zoomeye}.

Given an input image $I$ and question $q$, active perception can be abstracted as a sequence of task-conditioned perceptual actions:
\begin{equation}
    a_t^{p} \sim \pi_\theta(\cdot \mid I, q, h_{t-1}), 
    \qquad
    h_t = h_{t-1} \circ o_t ,
\end{equation}
where $a_t^{p}$ is the perceptual action at step $t$, $o_t$ is the acquired observation, and $h_t$ is the accumulated perception history.
In \mangavqa{}, the perceptual action is naturally defined over panels.
We therefore instantiate active perception as question-guided panel inspection, where each action selects one panel and updates a compact evidence sketch.

\paragraph{Group-Relative Policy Optimization.}
Reinforcement learning has become a common post-training strategy for improving LLM reasoning~\citep{yu2025dapo, gao2025soft, zheng2025group, shrivastava2025sample}.
Group Relative Policy Optimization (GRPO) is a critic-free variant of PPO that estimates advantages by comparing multiple responses sampled for the same prompt, avoiding the need to train a separate value model~\citep{shao2024deepseekmath}.
Given an input $x$, GRPO samples a group of responses $\{Y_i\}_{i=1}^{G}$ from the old policy and computes a normalized advantage from their rewards:
\begin{equation}
    Y_i \sim \pi_{\theta_{\mathrm{old}}}(\cdot \mid x),
    \qquad
    A_i = \frac{R_i-\mu_R}{\sigma_R+\delta},
\end{equation}
where $R_i$ is the reward of response $Y_i$, and $\mu_R,\sigma_R$ are the mean and standard deviation of rewards within the group.

The policy is then updated with a PPO-style clipped objective over generated tokens:
\begin{equation}
    \mathcal{L}_{\mathrm{GRPO}}
    =
    -\mathbb{E}_{i,t}
    \left[
    \min\left(
    \rho_{i,t}A_i,
    \bar{\rho}_{i,t}A_i
    \right)
    \right],
\end{equation}
where $\rho_{i,t}$ is the token-level probability ratio between the current and old policies, and $\bar{\rho}_{i,t}$ is its clipped version.
GRPO is well suited to our setting because \mangavqa{} naturally permits multiple sampled reasoning trajectories for the same question.
We therefore use group-relative optimization with rewards derived from sampled answers and their panel-level evidence paths, rather than from annotated answers or rationales.

\section{\method}
\label{sec:method}

We propose \method{}, an unsupervised reasoning framework for manga visual question answering that learns to answer questions by actively sketching question-relevant evidence across panels, as shown in~\Cref{fig:overview}.
It consists of Active Narrative Sketching for panel-level evidence construction and group-relative training for unsupervised policy optimization.
Given a manga page $I$, panel boxes $B=\{b_m\}_{m=1}^{M}$, and a question $q$, the policy model $\pi_\theta$ generates a structured sketch and a final answer.

\subsection{Active Narrative Sketching}
\label{sec:narrative_sot}

Manga pages present visual and textual evidence as panel-based narratives, where answering a question often requires selectively following characters, dialogue, actions, and cross-panel relations.
To model this process, we introduce Active Narrative Sketching (\activesketch{}), an active sketching procedure that decides \emph{where to look}, \emph{what to retain}, and \emph{when to stop}, yielding a compact panel-grounded sketch before answer generation.
\activesketch{} therefore connects perception and reasoning through a recurrent sketch state: each perception step adds one grounded clue, and the accumulated sketch serves as the evidence state for the final answer.

\paragraph{Active Panel Perception.}
Instead of passively answering from the whole page, \activesketch{} treats \mangavqa{} as an iterative evidence-seeking process over panels.
At each step, the policy conditions on the question, the manga page, and the current sketch state to perform three actions: 
(i) \emph{panel selection}, which chooses a panel $p_t \in \{1,\ldots,M\}$ as the next visual narrative unit to inspect; 
(ii) \emph{evidence sketching}, which extracts a concise clue $c_t$ that is visually supported by the selected panel and relevant to the question; and 
(iii) \emph{stopping decision}, which determines whether the accumulated sketch is sufficient for answering or another panel should be inspected.
This active process adapts visual computation to the question, rather than treating all panels as an unordered set of image regions.

Formally, the sketch state is initialized as $S_0=\emptyset$. 
At step $t$, the policy appends a panel-clue pair to the sketch:
\begin{equation}
   S_t = S_{t-1} \circ e_t, \qquad e_t=(p_t,c_t),
\end{equation}
where $p_t$ is the selected panel and $c_t$ is the extracted clue.
The process continues until the policy emits the stopping action, after which the accumulated sketch $S_T$ is passed to the answer generation stage.

\paragraph{Narrative Evidence Sketch.}
The step-wise updates above form a structured evidence sketch
\begin{equation}
    S = \big((p_1,c_1), (p_2,c_2), \ldots, (p_L,c_L)\big),
\end{equation}
where $p_t$ is the panel inspected at step $t$ and $c_t$ is a short clue grounded in that panel.
The sketch follows a question-directed evidence order rather than the page reading order: the model may first inspect a panel containing key dialogue, then move to another panel to identify the speaker, and later revisit a previous panel to verify a character relation.
Each clue is constrained to be \emph{local}, \emph{atomic}, and \emph{question-relevant}: it should describe evidence visible in the selected panel, focus on one useful fact, and contribute to answering the current question.

To make this active sketching process explicit and parseable, we implement \activesketch{} with a structured generation interface, where each step outputs a selected panel, a grounded clue, and a continue-or-answer decision:
\begin{center}
\begin{promptnewbox}{Active Sketching Prompt Template}
Given a manga page with visible panel IDs and a question $q$, construct a compact evidence sketch before answering.\\

\textbf{Step $t$:}\\
\textbf{Panel:} select one panel ID $p_t$ to inspect.\\
\textbf{Clue:} write one short clue $c_t$ visible in the selected panel and relevant to $q$.\\
\textbf{Decision:} choose \textsc{Continue} or \textsc{Answer}.\\

If \textsc{Answer}, generate the final answer from the collected clues.
\end{promptnewbox}
\end{center}

Once the model decides to answer, the final response is generated as
\begin{equation}
    \mathbf{a} \sim \pi_\theta(\mathbf{a} \mid I, B, q, S_T),
\end{equation}
where $\mathbf{a}$ is the predicted answer sequence.
For open-ended VQA, $\mathbf{a}$ is a natural-language response; for multiple-choice VQA, it is selected from the candidate option.

\subsection{Unsupervised Reward Modeling}
\label{sec:reward_modeling}

We optimize \activesketch{} without using human-annotated answers or rationale paths as rewards.
For each manga page $I$, panel boxes $B$, and question $q$, we sample $N$ rollouts from the current policy:
\begin{equation}
    Y_i \sim \pi_\theta(\cdot \mid I,B,q), \qquad i=1,\ldots,N,
\end{equation}
where each rollout $Y_i$ contains an active sketch and a final answer.
From each rollout, we parse the final answer and the ordered panel trajectory
\begin{equation}
    z_i=(p_{i,1},p_{i,2},\ldots,p_{i,L_i}),
    \label{eq:panel_trajectory}
\end{equation}
where $L_i$ is the number of perception steps in $Y_i$.
We then construct two complementary rewards: an answer preference reward that evaluates the sampled answer, and a path consistency reward that evaluates whether the panel trajectory is supported by stable evidence paths among sampled rollouts.

\paragraph{Answer Preference Reward.}
Since Manga VQA often admits open-ended answers, answer frequency or surface-form agreement may not reliably indicate answer quality.
We therefore estimate answer preference through listwise self-ranking.
For each valid rollout, we extract its final answer and merge duplicate answers into candidate clusters
\begin{equation}
    \mathcal{A}=\{\alpha_1,\ldots,\alpha_K\},
\end{equation}
where each $\alpha_k$ denotes a unique answer and may be produced by multiple rollouts.
Instances with $K<2$ are skipped since no relative preference can be formed.
We then ask the policy to rank the candidates under the same manga page and question, yielding a relative quality signal without annotated labels.
With $r_t^k$ denoting the rank position of $\alpha_k$ in the $t$-th sampled ranking, we compute
\begin{equation}
\begin{aligned}
    R_{\mathrm{pref}}(\alpha_k)
    &=
    \frac{1}{T(K-1)}
    \sum_{t=1}^{T}
    (K-r_t^k), \\
    R_{\mathrm{pref}}(Y_i)
    &= R_{\mathrm{pref}}(\alpha(Y_i)),
\end{aligned}
\end{equation}
where $\alpha(Y_i)$ is the answer cluster assigned to rollout $Y_i$.
The normalized score lies in $[0,1]$ and serves as the answer-level reward.

\paragraph{Path Consistency Reward.}
The answer reward scores the final response but does not directly evaluate whether the supporting panels form a stable reasoning path.
We therefore introduce a path consistency reward over the parsed panel trajectories.
Using the answer preference scores above, we select high-preference rollouts as reference paths and organize their trajectories into a prefix tree.
Let $\mathcal{H}$ denote rollouts with $R_{\mathrm{pref}}(Y_i)\geq\tau$.
For a prefix $u=(p_1,\ldots,p_\ell)$, its value is the average preference score of reference rollouts that contain $u$.
The consistency reward of rollout $Y_i$ is the average value of the prefixes along its own trajectory:
\begin{equation}
\begin{aligned}
    V(u)
    &=
    \frac{
    \sum_{j\in\mathcal{H}}
    \mathbb{I}[u \preceq z_j] R_{\mathrm{pref}}(Y_j)
    }{
    \sum_{j\in\mathcal{H}}
    \mathbb{I}[u \preceq z_j] + \epsilon
    }, \\
    R_{\mathrm{cons}}(Y_i)
    &=
    \frac{1}{L_i}
    \sum_{\ell=1}^{L_i}
    V(p_{i,1},\ldots,p_{i,\ell}),
\end{aligned}
\end{equation}
where $u \preceq z_j$ indicates that $u$ is a prefix of trajectory $z_j$.
Unlike set overlap, this reward is order-sensitive, allowing trajectories with the same panels but different orders to receive different scores.
This is aligned with \mangavqa{}, where answers often depend on ordered panels and cross-panel temporal or causal relations.

\subsection{Group-Relative Training}
\label{sec:training}

During training, we randomly permute visible panel IDs while keeping panel boxes fixed to avoid ID-order shortcuts, and then optimize the policy using the rewards derived above.

\paragraph{Reward Composition.}
For each rollout, we combine answer quality and path consistency into a single training reward:
\begin{equation}
\begin{aligned}
    R(y_i)
    =
    R_{\mathrm{fmt}}(y_i)
    \big(
    &\lambda_{\mathrm{pref}} R_{\mathrm{pref}}(y_i) \\
    &+
    \lambda_{\mathrm{cons}} R_{\mathrm{cons}}(y_i)
    \big),
\end{aligned}
\end{equation}
where $R_{\mathrm{fmt}}(y_i)\in\{0,1\}$ checks whether the rollout follows the required \activesketch{} format, and $\lambda_{\mathrm{pref}}$ and $\lambda_{\mathrm{cons}}$ balance the answer-quality and path-consistency terms.
The format reward prevents malformed outputs from receiving reward through accidental answer matches or panel overlaps.
Thus, a rollout is favored only when it produces a plausible answer and supports it with a stable panel-level evidence path.

\paragraph{Policy Optimization.}
We optimize the policy with group-relative policy optimization. 
For each question, sampled rollouts form a group, and rollout rewards are normalized into advantages:
\begin{equation}
    A_i =
    \frac{R(y_i)-\mu_R}{\sigma_R+\epsilon}.
\end{equation}
Let $Y_i=(y_{i,1},\ldots,y_{i,|Y_i|})$ denote the token sequence of rollout $i$. With the token-level ratio
\begin{equation}
    \rho_{i,t}(\theta)
    =
    \frac{
    \pi_\theta(y_{i,t}\mid y_{i,<t},I,B,q)
    }{
    \pi_{\theta_{\mathrm{old}}}(y_{i,t}\mid y_{i,<t},I,B,q)
    },
\end{equation}
we minimize
\begin{equation}
    \mathcal{L}_{\mathrm{RL}}(\theta)
    =
    -\mathbb{E}_{i,t}
    \left[
    \min\left(
    \rho_{i,t}A_i,
    \bar{\rho}_{i,t}A_i
    \right)
    \right],
\end{equation}
where $\bar{\rho}_{i,t}=\mathrm{clip}(\rho_{i,t},1-\epsilon,1+\epsilon)$.
The self-ranking samples are used only for reward construction and are not directly optimized, avoiding reinforcement of the model's own judging behavior.

\begin{table*}[!t]
    \centering
    \resizebox{\textwidth}{!}{
    \begin{tabular}{lcccccccc}
        \toprule
        \multirow{2}{*}{\textbf{Models}} 
        & \multicolumn{3}{c}{\textbf{MangaVQA}} 
        & \multicolumn{2}{c}{\textbf{ChrOMIC}} 
        & \multicolumn{3}{c}{\textbf{MangaUB}} \\
        \cmidrule(lr){2-4}\cmidrule(lr){5-6}\cmidrule(lr){7-9}
        & \textbf{Extraction} 
        & \textbf{Description} 
        & \textbf{Overall} 
        & \textbf{Multiple-choice} 
        & \textbf{Open-ended} 
        & \textbf{Single-panel} 
        & \textbf{Multi-panel} 
        & \textbf{Overall} \\
        \midrule
        \rowcolor{sectiongray}
        \multicolumn{9}{c}{\textit{General Perception}} \\
        \midrule
        MangaLMM$^\ddagger$ & 6.87 & 6.38 & 6.60 & 0.5108 & 3.83 & 0.6307 & 0.3958 & 0.5065 \\
        Molmo2-8B & 2.71 & 2.86 & 2.79 & 0.4372 & 4.89 & 0.7547 & \underline{0.5358} & \underline{0.6389} \\
        InternVL3.5-8B & 3.40 & 3.11 & 3.24 & 0.4978 & 5.40 & 0.7571 & 0.5092 & 0.6260 \\
        MiniCPM-V 4.5 & 4.41 & 4.40 & 4.24 & 0.5541 & 5.18 & \underline{0.7647} & 0.4890 & 0.6189 \\
        Keye-VL-1.5-8B & 3.57 & 2.65 & 3.07 & 0.4459 & 4.15 & 0.7624 & 0.4590 & 0.6020 \\
        LLaVA-OneVision-1.5 & 2.96 & 3.15 & 3.07 & 0.3030 & 5.34 & 0.7447 & 0.3741 & 0.5488 \\
        Kimi-VL-A3B-Instruct & 4.32 & 3.51 & 3.88 & 0.3983 & 5.53 & 0.7530 & 0.4413 & 0.5882 \\
        Qwen2.5-VL-7B-Instruct & 5.88 & 4.82 & 5.31 & 0.5298 & 5.54 & 0.7156 & 0.3826 & 0.5388 \\
        Qwen3-VL-8B-Instruct & \underline{7.18} & 5.74 & 6.40 & 0.4242 & 5.13 & 0.7568 & 0.4818 & 0.6114 \\
        \midrule
        \rowcolor{sectiongray}
        \multicolumn{9}{c}{\textit{Active Perception}} \\
        \midrule
        TreeVGR & 4.41 & 3.09 & 3.69 & 0.4719 & 2.82 & 0.7005 & 0.4089 & 0.5464 \\
        DeepEyes & 5.00 & 4.51 & 4.73 & 0.5195 & 5.67 & 0.7303 & 0.3984 & 0.5548 \\
        ViGoRL & 4.33 & 3.72 & 3.99 & 0.4935 & 5.38 & -- & -- & -- \\
        \midrule
        \rowcolor{oursblue}
        \method{} (Qwen2.5) & 6.16 & 5.45 & 5.78 & 0.5368 & 5.60 & 0.7248 & 0.4197 & 0.5635 \\
        \rowcolor{oursblue}
        \method{} (Qwen2.5)$^\ddagger$ & 7.05 & \underline{6.47} & \underline{6.73} & \underline{0.5671} & \underline{6.06} & 0.7584 & 0.4535 & 0.5972 \\
        \rowcolor{oursblue}
        \method{} (Qwen3) & \textbf{7.20} & \textbf{6.64} & \textbf{6.90} & \textbf{0.5801} & \textbf{6.59} & \textbf{0.7711} & \textbf{0.5536} & \textbf{0.6561} \\
        \bottomrule
    \end{tabular}
    }
    \caption{
        Main results on three manga understanding benchmarks. MangaVQA is evaluated on \textit{Exact Extraction}, \textit{Descriptive Answering}, and overall performance; ChrOMIC is evaluated on \textit{Multiple-choice} and \textit{Open-ended} settings; and MangaUB is evaluated on \textit{Single-panel}, \textit{Multi-panel}, and overall performance. Models marked with $^\ddagger$ are fine-tuned on MangaOCR data. Best results are in bold and second-best results are underlined.
    }
    \label{tab:main}
\end{table*}

\section{Experiments}

\subsection{Experimental Settings}
\label{sec:settings}

\paragraph{Datasets and Benchmarks.}


For unsupervised training, we construct a question-only subset from the training set of MangaVQA~\citep{baek2026mangavqa}. 
We first remove samples with empty questions/answers or identical question-answer texts and then randomly sample 512 questions, using only the questions and their corresponding Manga109 panel annotations~\citep{aizawa2020building}; answer labels are not used for reward construction.
For evaluation, we use three manga understanding benchmarks. 
(1) \textbf{MangaVQA} evaluates open-ended \mangavqa{} with \textit{Exact Extraction} and \textit{Descriptive Answering} subsets, and we report both subset and overall scores. 
(2) \textbf{ChrOMIC}~\citep{hou2026chromic} evaluates chronological reasoning over comics; we use its Japanese manga subset and report both the original multiple-choice setting and a manually converted open-ended setting. For the latter, we retain only questions that admit free-form answers and use the text of the original correct option as the reference answer. 
(3) \textbf{MangaUB}~\citep{ikuta2025mangaub} is a large-scale multiple-choice benchmark for manga understanding. We report results separately for single- and multi-panel questions, which respectively probe within-panel evidence extraction and cross-panel evidence aggregation, along with overall performance.

\paragraph{Evaluation Protocol.} 
Following the official MangaVQA evaluation protocol, we employ Gemini 2.5 Flash~\citep{comanici2025gemini} (\texttt{gemini-2.5-flash}) as an LLM judge for open-ended evaluation. Given a question, a reference answer, and a model-generated response, the judge assesses the appropriateness and relevance of the response with respect to the reference and assigns a score from 1 to 10. We apply the same protocol to the open-ended ChrOMIC setting. For the multiple-choice ChrOMIC setting and the MangaUB evaluations, we report accuracy. Unless otherwise specified, all results are averaged over three sampled responses per question.

\paragraph{Baselines.}
We compare \method{} with two baselines. 
(a) \textbf{General Perception} covers manga-specialized and general-purpose MLLMs~\citep{baek2026mangavqa, clark2026molmo2, wang2025internvl3, yu2025minicpm, yang2025kwai, an2025llava, team2025kimi, bai2025qwen25vltechnicalreport, bai2025qwen3} that answer from a single-pass perception of the full page, without explicitly searching for intermediate visual evidence. 
(b) \textbf{Active Perception} includes methods that iteratively inspect visual regions or acquire visual evidence before prediction~\citep{wang2025traceable, zheng2025deepeyes, sarch2025grounded} which are based on Qwen2.5-VL.

\paragraph{Implementation Details.}
We use Qwen3-VL-8B-Instruct~\citep{bai2025qwen3} as the base model and train it with \method{} on two NVIDIA A800 GPUs for 11 epochs, with a total training time of approximately 18 hours. 
We set the batch size to 8 and the learning rate to $4\times10^{-6}$. 
For each training instance, we sample 8 rollouts from the policy and query the policy 8 times for answer ranking. 
For reward modeling, the preference threshold is $\tau=0.5$, and the reward weights are $\lambda_{\mathrm{pref}}=0.7$ and $\lambda_{\mathrm{cons}}=0.3$. 
Following recent observations that intrinsic unsupervised RL can become unstable when scaled aggressively~\citep{he2026far}, we use the compact training subset described above.
During training, visible panel IDs are randomly permuted while panel boxes remain fixed to reduce ID-order shortcuts.

\begin{table}[t]
    \centering
    \resizebox{\columnwidth}{!}{
        \begin{tabular}{cccccc}
        \toprule
        \multirow{2}{*}{\textbf{ANS}}
        & \multirow{2}{*}{$\boldsymbol{R_{\mathrm{pref}}}$}
        & \multirow{2}{*}{$\boldsymbol{R_{\mathrm{cons}}}$}
        & \multirow{2}{*}{\textbf{MangaVQA}} 
        & \multicolumn{2}{c}{\textbf{ChrOMIC}} \\
        \cmidrule(lr){5-6}
        & & & & \textbf{MC} & \textbf{OE} \\
        \midrule    
        \cmark & \xmark & \xmark & 6.45 & 0.4285 & 5.18 \\
        \xmark & \cmark & \xmark & 6.60 & 0.4372 & 6.12 \\
        \cmark & \cmark & \xmark & 6.67 & 0.4545 & 6.31 \\       
        \cmark & \cmark & \cmark & \textbf{6.90} & \textbf{0.5801} & \textbf{6.59} \\
        \bottomrule
        \end{tabular}
    }
    \caption{
        Ablation study on the main components of \method{}. 
        $R_{\mathrm{pref}}$ denotes the answer preference reward, and $R_{\mathrm{cons}}$ denotes the path consistency reward. 
    }
    \label{tab:component_ablation}
\end{table}

\subsection{Main Results}
\label{sec:main_results}

\begin{figure*}[t]
    \centering
    \includegraphics[width=\linewidth]{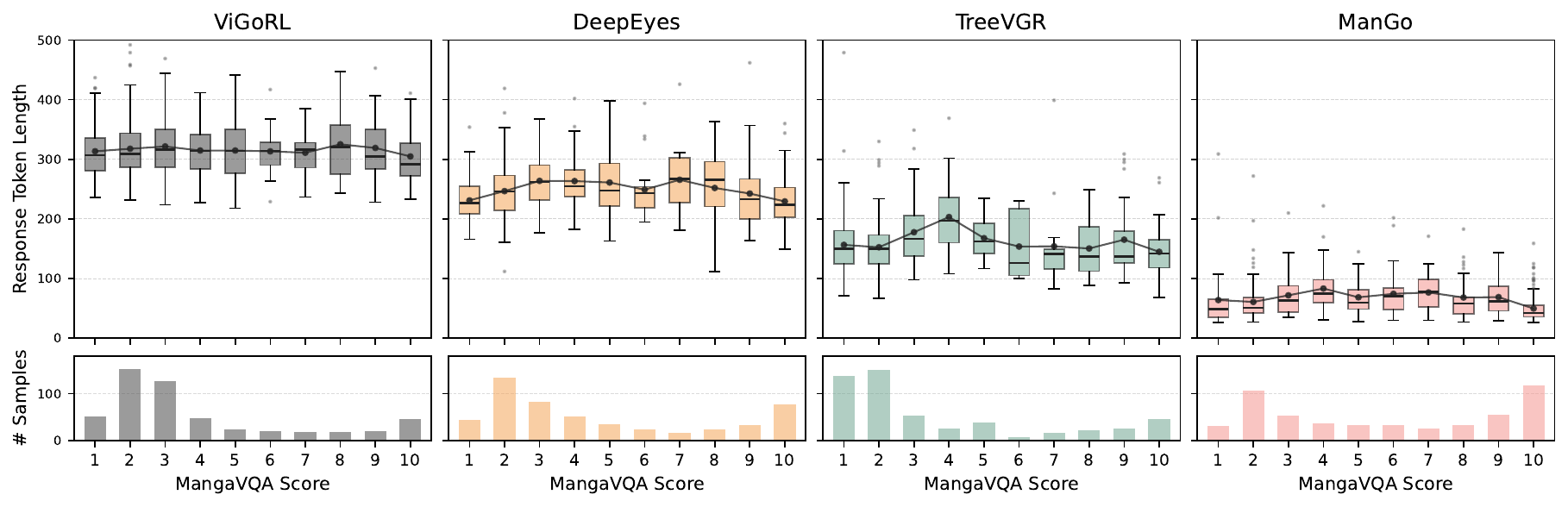}
    \caption{
        Response analysis on MangaVQA. Boxplots show the distribution of response token lengths within each score bin, and lower bars indicate the number of samples. Compared with other active perception methods, \method{} concentrates more samples in high scores while maintaining shorter responses.
    }
    \label{fig:token_length}
\end{figure*}

\Cref{tab:main} compares \method{} with general perception and active perception baselines across three manga understanding benchmarks.
Overall, \method{} achieves the strongest and most consistent performance across open-ended, multiple-choice, single-panel, and multi-panel settings.
Compared with general perception models, \method{} benefits from explicitly decomposing manga VQA into panel-level evidence construction before answer generation, which is better aligned with the narrative structure of manga pages.
Compared with active perception baselines, \method{} does not merely inspect visual regions; it retains selected evidence as a compact narrative sketch and updates the sketch through question-guided panel perception.
These results indicate that \method{} is more effective on tasks that require linking characters, dialogue, and events across panels.

Since MangaLMM incorporates MangaOCR supervision in addition to MangaVQA training, we also apply \method{} to a Qwen2.5-VL checkpoint that has been fine-tuned on MangaOCR under the same SFT protocol as MangaLMM. Combined with the improvements over the vanilla Qwen2.5-VL and Qwen3-VL baselines reported earlier, these results demonstrate that \method{} remains effective across different backbones and initializations.

\subsection{Ablation Study}

\paragraph{Component Ablation.}
\Cref{tab:component_ablation} validates the contribution of each component.
\activesketch{} alone provides a structured panel-grounded reasoning format, but without reward optimization its performance remains limited.
$R_{\mathrm{pref}}$ improves answer quality, especially for open-ended evaluation, while combining it with \activesketch{} yields more stable gains by optimizing answers within a structured sketch.
Adding $R_{\mathrm{cons}}$ further improves all settings, showing that path consistency complements answer preference by encouraging high-quality answers to follow stable panel trajectories.
The full model performs best, confirming that active sketching, answer preference, and path consistency are beneficial.

\paragraph{Preference Reward Ablation.}
\Cref{tab:pref_ablation} compares answer preference estimators with \activesketch{}, path consistency, and policy optimization fixed.
Clustering~\citep{zhang2025right} or confidence~\citep{prabhudesai2025maximizing} based rewards provide weak supervision for manga VQA, where answer quality depends on fine-grained speaker, character, and event relations.
POLAR~\cite{dou2025pre} remains competitive but underperforms our self-ranking reward despite using ground-truth answers, indicating that reference matching can miss narrative distinctions.
Self-ranking is more effective because it evaluates candidate answers relative to other rollouts under the same page and question, which better matches the group-relative training objective.

\begin{table}[t]
    \centering
    \resizebox{\columnwidth}{!}{
        \begin{tabular}{lcccc}
        \toprule
        \multirow{2}{*}{\textbf{Preference Reward}} 
        & \multirow{2}{*}{\textbf{GT}}
        & \multirow{2}{*}{\textbf{MangaVQA}}
        & \multicolumn{2}{c}{\textbf{ChrOMIC}} \\
        \cmidrule(lr){4-5}
        & & & \textbf{MC} & \textbf{OE} \\
        \midrule
        POLAR & \cmark & 6.35 & 0.5411 & 5.91 \\
        EMPO & \xmark & 4.75 & 0.4978 & 2.67 \\
        RENT & \xmark & 1.01 & 0.2554 & 1.00 \\
        Self-Ranking (ours) & \xmark & \textbf{6.90} & \textbf{0.5801} & \textbf{6.59} \\
        \bottomrule
    \end{tabular}
    }
    \caption{
        Ablation study on preference reward estimators, with \activesketch{}, path consistency, and policy optimization fixed.
        \textit{GT} indicates whether ground-truth answers are used in reward computation.
    }
    \label{tab:pref_ablation}
\end{table}

\paragraph{Consistency Reward Ablation.}
\Cref{tab:cons_ablation} compares different estimators for the path consistency reward while keeping \activesketch{}, answer preference reward, and policy optimization fixed.
Jaccard matching performs bad because it ignores panel order and only measures unordered overlap.
Sequence-based metrics improve by considering ordered panel paths, showing that the structure of the evidence trajectory matters for manga.
Prefix Tree performs best across all settings, suggesting that rewarding stable prefixes better captures how early panel choices guide subsequent evidence construction.
This supports our design of path consistency reward, which stabilizes question-directed panel trajectories rather than merely matching selected panels.

\begin{table}[t]
    \centering
    \resizebox{\columnwidth}{!}{
        \begin{tabular}{lccc}
        \toprule
        \multirow{2}{*}{\textbf{Consistency Reward}} 
        & \multirow{2}{*}{\textbf{MangaVQA}}
        & \multicolumn{2}{c}{\textbf{ChrOMIC}} \\
        \cmidrule(lr){3-4}
        & & \textbf{MC} & \textbf{OE} \\
        \midrule
        Jaccard & 6.70 & 0.4762 & 6.33 \\
        Longest Common Subsequence & 6.76 & 0.5108 & 6.41 \\
        Longest Common Substring & 6.79 & 0.5325 & 6.47 \\
        Prefix Tree (ours) & \textbf{6.90} & \textbf{0.5801} & \textbf{6.59} \\
        \bottomrule
    \end{tabular}
    }
    \caption{
        Ablation study on consistency reward estimators.
        All variants use the same \activesketch{} generation, answer preference reward, and policy optimization, differing only in how path consistency is computed.
    }
    \label{tab:cons_ablation}
\end{table}

\begin{figure*}[t]
    \centering
    \includegraphics[width=\linewidth]{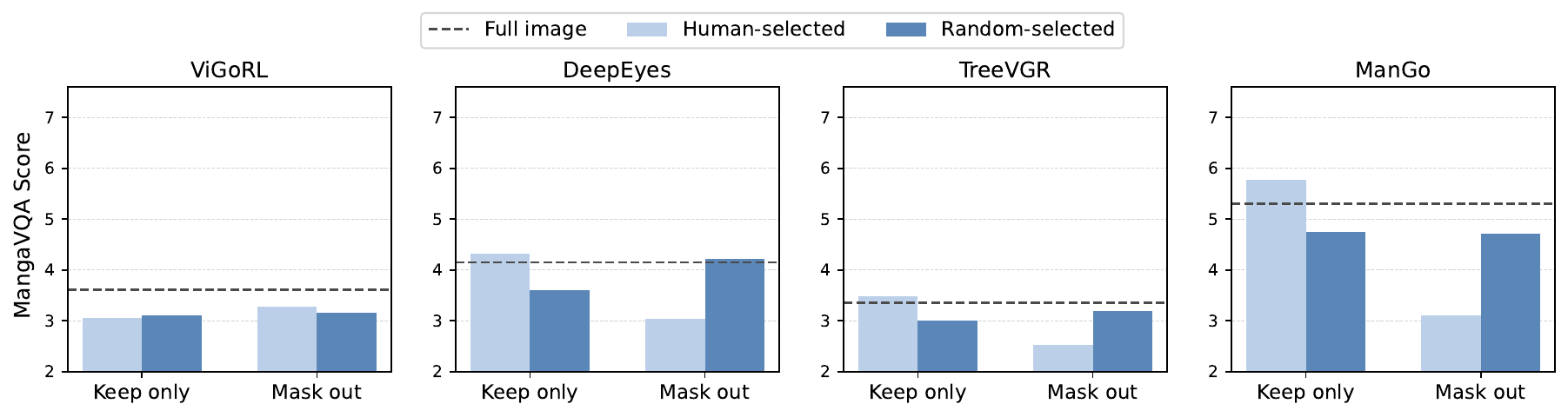}
    \caption{
        Panel perturbation analysis on MangaVQA. Each subplot compares manually annotated answer-supporting panels with size-matched random panels under keep-only and mask-out settings. 
    }
    \label{fig:mask_all}
\end{figure*}

\subsection{Analysis}
\label{sec:analysis}
\begin{figure}[t]
    \centering
    \includegraphics[width=\linewidth]{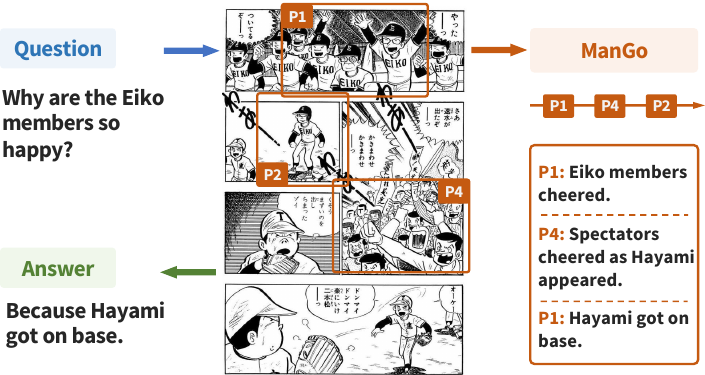}
    \caption{
        Case study of \method{}.
        Given a question about why the Eiko members are happy, \method{} selects supporting panels, sketches concise clues about the team's reaction and Hayami's action, and answers by linking the emotion to its narrative cause. \copyright Mikio Yoshimori
    }
    \label{fig:case}
\end{figure}

\paragraph{Evaluation Robustness.}

To examine whether the open-ended evaluation is sensitive to the choice of LLM judge, we rescored the same MangaVQA responses using GPT-4o~\citep{hurst2024gpt} as an alternative to Gemini 2.5 Flash. As shown in \Cref{tab:robustness}, although GPT-4o generally assigns lower absolute scores, the overall relative performance remains consistent, with \method{} achieving the highest score under both judges. These results support the reliability of the evaluation across judge models.

\begin{table}[!]
\centering
\resizebox{\columnwidth}{!}{
\begin{tabular}{lcc}
\toprule
\textbf{Models} & \textbf{Judge: Gemini} & \textbf{Judge: GPT-4o} \\
\midrule
MangaLMM & 6.68 & 6.57 \\
Molmo2-8B & 2.79 & 2.15 \\
InternVL3.5-8B & 3.24 & 3.02 \\
MiniCPM-V 4.5 & 4.24 & 3.98 \\
Keye-VL-1.5-8B & 3.07 & 2.38 \\
LLaVA-OneVision-1.5 & 3.07 & 2.44 \\
Kimi-VL-A3B-Instruct & 3.88 & 3.52 \\
Qwen2.5-VL-7B-Instruct & 5.31 & 4.92 \\
Qwen3-VL-8B-Instruct & 6.40 & 6.03 \\
TreeVGR & 3.69 & 3.58 \\
DeepEyes & 4.73 & 4.55 \\
ViGoRL & 3.99 & 3.81 \\
\midrule
\rowcolor{oursblue}
\method{} & \textbf{6.90} & \textbf{6.78} \\
\bottomrule
\end{tabular}
}
\caption{Robustness of LLM-as-a-judge evaluation on MangaVQA. The same model responses are scored by Gemini 2.5 Flash and GPT-4o respectively.}
\label{tab:robustness}
\end{table}

\paragraph{Response Compactness.}
\Cref{fig:token_length} shows that existing active perception methods tend to produce substantially longer responses, suggesting that iterative visual exploration can introduce redundant or weakly organized information.
In contrast, \method{} maintains a compact response distribution while achieving stronger MangaVQA performance.
This supports the design of Active Narrative Sketching: instead of accumulating lengthy visual descriptions, the model records local, atomic, and question-relevant clues in a structured sketch.
The result suggests that manga VQA benefits more from concise evidence organization than from expanding the amount of intermediate text.

\paragraph{Evidence Utilization.}


\Cref{fig:mask_all} evaluates whether models rely on panels that actually support the answer, using keep-only and mask-out perturbations. Answer-supporting panels are manually annotated under a necessity criterion: a panel is labeled as supporting if its removal precludes derivation of the correct answer. Size-matched randomly selected panels are used as controls.
\method{} benefits more when human-selected panels are retained and drops more when they are masked out, while random-selected panels lead to weaker or less consistent effects.
This indicates that \method{} is more tightly grounded in answer-relevant panels, rather than relying on incidental context or generic visual exploration.
This result is consistent with our reward design: path consistency encourages high-quality answers to be supported by stable ordered panel trajectories, making the learned policy more sensitive to answer-supporting panels.

\paragraph{Case Study.}
\Cref{fig:case} shows a representative reasoning path produced by \method{}.
Given the question about why the Eiko members are happy, the model selects panels showing the cheering team, Hayami's appearance, and Hayami getting on base, then composes these clues into a compact sketch.
The final answer follows from this panel-level narrative link rather than from a single salient visual cue, illustrating how \method{} grounds manga reasoning in ordered supporting evidence.

\section{Related Work}

\paragraph{Visual Question Answering.} VQA has evolved from object- and attribute-centric question answering on natural images toward more complex multimodal understanding scenarios. Recent studies increasingly examine models in settings that involve text-rich scenes, multilingual visual content, scientific figures and ambiguous visual contexts~\citep{tang2025mtvqa, jiu2025tvqacml, burgess2025microvqa, jian2025teaching, yang2025magic, xiao2025egoblind}. This trend suggests that VQA is shifting from simple visual recognition to reasoning over structured visual information and task-specific contexts, of which \mangavqa{} is a representative but less explored case.

\paragraph{Manga VQA.} Existing manga and comic benchmarks have studied related abilities such as reading order, panel sequencing, speaker identification, and multi-panel understanding~\citep{vivoli2024comix, ikuta2025mangaub, vivoli2025comicspap, wang2025beyond, hou2026chromic}. MangaVQA and MangaLMM introduce a manga-specific VQA benchmark and specialized model for manga understanding~\citep{baek2026mangavqa}. Nevertheless, compared with general VQA and other specialized VQA domains, \mangavqa{} remains relatively underexplored, especially for open-ended question answering over panel-structured narratives.

\section{Conclusion}
\label{sec:conclusion}

We present \method{}, an unsupervised framework for manga visual question answering that treats answering as active panel-level evidence seeking. 
Through Active Narrative Sketching, the model selects relevant panels, extracts concise grounded clues, and builds a compact evidence sketch before answering. 
With answer preference and path consistency rewards optimized by group-relative training, \method{} learns this behavior without human-annotated answers or rationale paths. 
Experiments on multiple manga understanding benchmarks show that \method{} improves both final answer accuracy and evidence grounding.

\section*{Limitations}

Although \method{} achieves strong performance on existing manga understanding benchmarks, its evaluation is still constrained by the limited availability of large-scale manga VQA resources. Current benchmarks cover only a portion of the broad spectrum of manga comprehension, with data sources that remain limited in genre, era, language, and cultural context. Therefore, the reported results may not fully characterize model behavior across more diverse manga scenarios. Future work could develop broader benchmarks and evaluation protocols to support a more comprehensive assessment of manga understanding systems.

\section*{Ethics Statement}

This work studies \mangavqa{} using existing manga understanding benchmarks for academic research. Manga pages are copyrighted creative works, and we follow the usage conditions of the corresponding datasets. We do not redistribute raw manga images beyond what is permitted by the original resources. 
All manual annotations used in this work were conducted by the authors and disagreements are resolved through majority voting. No external annotators, crowdworkers, or human-subject participants were recruited, and no annotator personal data were collected.
Since manga may contain culturally specific expressions, sensitive themes, or subjective narrative interpretations, model predictions should not be treated as authoritative readings of the original works. We encourage future use of this work to respect dataset licenses, avoid unauthorized redistribution of copyrighted materials, and conduct additional human verification when applying \mangavqa{} systems in user-facing scenarios.

\section*{Acknowledgement}

This work was supported in part by the National Natural Science Foundation of China under Grant No. 62332010 and by the Fundamental Research Funds for the Central Universities under Grant No. CUC26TD06.

\bibliography{custom}

\begin{thebibliography}{38}
\providecommand{\natexlab}[1]{#1}

\bibitem[{Aizawa et~al.(2020)Aizawa, Fujimoto, Otsubo, Ogawa, Matsui, Tsubota,
  and Ikuta}]{aizawa2020building}
Kiyoharu Aizawa, Azuma Fujimoto, Atsushi Otsubo, Toru Ogawa, Yusuke Matsui,
  Koki Tsubota, and Hikaru Ikuta. 2020.
\newblock Building a manga dataset “manga109” with annotations for
  multimedia applications.
\newblock \emph{IEEE multimedia}, 27(2):8--18.

\bibitem[{Aloimonos et~al.(1988)Aloimonos, Weiss, and
  Bandyopadhyay}]{aloimonos1988active}
John Aloimonos, Isaac Weiss, and Amit Bandyopadhyay. 1988.
\newblock Active vision.
\newblock \emph{International journal of computer vision}, 1(4):333--356.

\bibitem[{An et~al.(2025)An, Xie, Yang, Zhang, Zhao, Cheng, Wang, Xu, Chen, Zhu
  et~al.}]{an2025llava}
Xiang An, Yin Xie, Kaicheng Yang, Wenkang Zhang, Xiuwei Zhao, Zheng Cheng,
  Yirui Wang, Songcen Xu, Changrui Chen, Didi Zhu, et~al. 2025.
\newblock Llava-onevision-1.5: Fully open framework for democratized multimodal
  training.
\newblock \emph{arXiv preprint arXiv:2509.23661}.

\bibitem[{Baek et~al.(2026)Baek, Egashira, Onohara, Miyai, Imajuku, Ikuta, and
  Aizawa}]{baek2026mangavqa}
Jeonghun Baek, Kazuki Egashira, Shota Onohara, Atsuyuki Miyai, Yuki Imajuku,
  Hikaru Ikuta, and Kiyoharu Aizawa. 2026.
\newblock Mangavqa and mangalmm: A benchmark and specialized model for
  multimodal manga understanding.
\newblock In \emph{Findings of the Association for Computational Linguistics:
  EACL 2026}, pages 5349--5370.

\bibitem[{Bai et~al.(2025{\natexlab{a}})Bai, Cai, Chen, Chen, Chen, Cheng,
  Deng, Ding, Gao, Ge et~al.}]{bai2025qwen3}
Shuai Bai, Yuxuan Cai, Ruizhe Chen, Keqin Chen, Xionghui Chen, Zesen Cheng,
  Lianghao Deng, Wei Ding, Chang Gao, Chunjiang Ge, et~al. 2025{\natexlab{a}}.
\newblock Qwen3-vl technical report.
\newblock \emph{arXiv preprint arXiv:2511.21631}.

\bibitem[{Bai et~al.(2025{\natexlab{b}})Bai, Chen, Liu, Wang, Ge, Song, Dang,
  Wang, Wang, Tang, Zhong, Zhu, Yang, Li, Wan, Wang, Ding, Fu, Xu, Ye, Zhang,
  Xie, Cheng, Zhang, Yang, Xu, and Lin}]{bai2025qwen25vltechnicalreport}
Shuai Bai, Keqin Chen, Xuejing Liu, Jialin Wang, Wenbin Ge, Sibo Song, Kai
  Dang, Peng Wang, Shijie Wang, Jun Tang, Humen Zhong, Yuanzhi Zhu, Mingkun
  Yang, Zhaohai Li, Jianqiang Wan, Pengfei Wang, Wei Ding, Zheren Fu, Yiheng
  Xu, and 8 others. 2025{\natexlab{b}}.
\newblock \href {https://arxiv.org/abs/2502.13923} {Qwen2.5-vl technical
  report}.
\newblock \emph{Preprint}, arXiv:2502.13923.

\bibitem[{Burgess et~al.(2025)Burgess, Nirschl, Bravo-S{\'a}nchez, Lozano,
  Gupte, Galaz-Montoya, Zhang, Su, Bhowmik, Coman et~al.}]{burgess2025microvqa}
James Burgess, Jeffrey~J Nirschl, Laura Bravo-S{\'a}nchez, Alejandro Lozano,
  Sanket~Rajan Gupte, Jesus~G Galaz-Montoya, Yuhui Zhang, Yuchang Su, Disha
  Bhowmik, Zachary Coman, et~al. 2025.
\newblock Microvqa: A multimodal reasoning benchmark for microscopy-based
  scientific research.
\newblock In \emph{Proceedings of the IEEE/CVF Conference on Computer Vision
  and Pattern Recognition}, pages 19552--19564.

\bibitem[{Clark et~al.(2026)Clark, Zhang, Ma, Park, Tripathi, Lee, Salehi, Ren,
  Kim, Yang et~al.}]{clark2026molmo2}
Christopher Clark, Jieyu Zhang, Zixian Ma, Jae~Sung Park, Rohun Tripathi,
  Sangho Lee, Mohammadreza Salehi, Jason Ren, Chris~Dongjoo Kim, Yinuo Yang,
  et~al. 2026.
\newblock Molmo2: Open weights and data for vision-language models with video
  understanding and grounding.
\newblock In \emph{Proceedings of the IEEE/CVF Conference on Computer Vision
  and Pattern Recognition}, pages 28652--28668.

\bibitem[{Comanici et~al.(2025)Comanici, Bieber, Schaekermann, Pasupat,
  Sachdeva, Dhillon, Blistein, Ram, Zhang, Rosen et~al.}]{comanici2025gemini}
Gheorghe Comanici, Eric Bieber, Mike Schaekermann, Ice Pasupat, Noveen
  Sachdeva, Inderjit Dhillon, Marcel Blistein, Ori Ram, Dan Zhang, Evan Rosen,
  et~al. 2025.
\newblock Gemini 2.5: Pushing the frontier with advanced reasoning,
  multimodality, long context, and next generation agentic capabilities.
\newblock \emph{arXiv preprint arXiv:2507.06261}.

\bibitem[{Dou et~al.(2025)Dou, Liu, Yang, Zou, Zhou, Xing, Huang, Ge, Lv, Song,
  Gao, Lyu, Zhou, Guo, Xi, Guo, Zhang, Gui, Zhang, Qiu, Huang, and
  Chen}]{dou2025pre}
Shihan Dou, Shichun Liu, Yuming Yang, Yicheng Zou, Yunhua Zhou, Shuhao Xing,
  Chenhao Huang, Qiming Ge, haijun Lv, Demin Song, Songyang Gao, Chengqi Lyu,
  Enyu Zhou, Honglin Guo, Zhiheng Xi, Qipeng Guo, Wenwei Zhang, Tao Gui,
  Qi~Zhang, and 3 others. 2025.
\newblock Pre-trained policy discriminators are general reward models.
\newblock In \emph{Advances in Neural Information Processing Systems}, volume
  38, Main Conference, pages 170177--170217. Curran Associates, Inc.

\bibitem[{Gao et~al.(2025)Gao, Zheng, Chen, Dang, Liu, Yu, Yang, Bai, Zhou, and
  Lin}]{gao2025soft}
Chang Gao, Chujie Zheng, Xiong-Hui Chen, Kai Dang, Shixuan Liu, Bowen Yu,
  An~Yang, Shuai Bai, Jingren Zhou, and Junyang Lin. 2025.
\newblock Soft adaptive policy optimization.
\newblock \emph{arXiv preprint arXiv:2511.20347}.

\bibitem[{He et~al.(2026)He, Zuo, Liu, Zhao, Fu, Yang, Qian, Zhang, Fan, Cui
  et~al.}]{he2026far}
Bingxiang He, Yuxin Zuo, Zeyuan Liu, Shangziqi Zhao, Zixuan Fu, Junlin Yang,
  Cheng Qian, Kaiyan Zhang, Yuchen Fan, Ganqu Cui, et~al. 2026.
\newblock How far can unsupervised rlvr scale llm training?
\newblock \emph{arXiv preprint arXiv:2603.08660}.

\bibitem[{Hou et~al.(2026)Hou, Lin, Zhang, Yin, Zhu, Hong, Gao, and
  Wang}]{hou2026chromic}
Bingxuan Hou, Jiayi Lin, Chenyang Zhang, Dapeng Yin, Shuyue Zhu, Qingqing Hong,
  Mengna Gao, and Junli Wang. 2026.
\newblock Chromic: Chronological reasoning across multi-panel comics.
\newblock In \emph{Proceedings of the 19th Conference of the European Chapter
  of the Association for Computational Linguistics (Volume 1: Long Papers)},
  pages 4384--4400.

\bibitem[{Hurst et~al.(2024)Hurst, Lerer, Goucher, Perelman, Ramesh, Clark,
  Ostrow, Welihinda, Hayes, Radford et~al.}]{hurst2024gpt}
Aaron Hurst, Adam Lerer, Adam~P Goucher, Adam Perelman, Aditya Ramesh, Aidan
  Clark, AJ~Ostrow, Akila Welihinda, Alan Hayes, Alec Radford, et~al. 2024.
\newblock Gpt-4o system card.
\newblock \emph{arXiv preprint arXiv:2410.21276}.

\bibitem[{Ikuta et~al.(2025)Ikuta, W{\"o}hler, and Aizawa}]{ikuta2025mangaub}
Hikaru Ikuta, Leslie W{\"o}hler, and Kiyoharu Aizawa. 2025.
\newblock Mangaub: A manga understanding benchmark for large multimodal models.
\newblock \emph{IEEE MultiMedia}, 32(2):33--43.

\bibitem[{Jian et~al.(2025)Jian, Yu, Yang, Ren, and Zhang}]{jian2025teaching}
Pu~Jian, Donglei Yu, Wen Yang, Shuo Ren, and Jiajun Zhang. 2025.
\newblock Teaching vision-language models to ask: Resolving ambiguity in visual
  questions.
\newblock In \emph{Proceedings of the 63rd Annual Meeting of the Association
  for Computational Linguistics (Volume 1: Long Papers)}, pages 3619--3638.

\bibitem[{Jiu et~al.(2025)Jiu, Weng, Zhu, Feng, Liu, and
  Jialedongzhu}]{jiu2025tvqacml}
Sha Jiu, Yu~Weng, Mengxiao Zhu, Chong Feng, Zheng Liu, and Jialedongzhu. 2025.
\newblock {TVQACML}: Benchmarking text-centric visual question answering in
  multilingual {C}hinese minority languages.
\newblock In \emph{Proceedings of the 2025 Conference on Empirical Methods in
  Natural Language Processing}, pages 13957--13967, Suzhou, China. Association
  for Computational Linguistics.

\bibitem[{Prabhudesai et~al.(2025)Prabhudesai, Chen, Ippoliti, Fragkiadaki,
  Liu, and Pathak}]{prabhudesai2025maximizing}
Mihir Prabhudesai, Lili Chen, Alex Ippoliti, Katerina Fragkiadaki, Hao Liu, and
  Deepak Pathak. 2025.
\newblock Maximizing confidence alone improves reasoning.
\newblock \emph{arXiv preprint arXiv:2505.22660}.

\bibitem[{Sarch et~al.(2025)Sarch, Saha, Khandelwal, Jain, Tarr, Kumar, and
  Fragkiadaki}]{sarch2025grounded}
Gabriel Sarch, Snigdha Saha, Naitik Khandelwal, Ayush Jain, Michael Tarr,
  Aviral Kumar, and Katerina Fragkiadaki. 2025.
\newblock Grounded reinforcement learning for visual reasoning.
\newblock In \emph{Advances in Neural Information Processing Systems}, volume
  38, Main Conference, pages 150977--151013. Curran Associates, Inc.

\bibitem[{Shao et~al.(2024)Shao, Wang, Zhu, Xu, Song, Bi, Zhang, Zhang, Li, Wu,
  and Guo}]{shao2024deepseekmath}
Zhihong Shao, Peiyi Wang, Qihao Zhu, Runxin Xu, Junxiao Song, Xiao Bi, Haowei
  Zhang, Mingchuan Zhang, Y.~K. Li, Y.~Wu, and Daya Guo. 2024.
\newblock \href {https://arxiv.org/abs/2402.03300} {Deepseekmath: Pushing the
  limits of mathematical reasoning in open language models}.
\newblock \emph{Preprint}, arXiv:2402.03300.

\bibitem[{Shen et~al.(2025)Shen, Zhao, Zhao, Xu, Zhang, Zhu, and
  Yin}]{shen2025zoomeye}
Haozhan Shen, Kangjia Zhao, Tiancheng Zhao, Ruochen Xu, Zilun Zhang, Mingwei
  Zhu, and Jianwei Yin. 2025.
\newblock {Z}oom{E}ye: Enhancing multimodal {LLM}s with human-like zooming
  capabilities through tree-based image exploration.
\newblock In \emph{Proceedings of the 2025 Conference on Empirical Methods in
  Natural Language Processing}, pages 6602--6618, Suzhou, China. Association
  for Computational Linguistics.

\bibitem[{Shrivastava et~al.(2026)Shrivastava, Awadallah, Balachandran, Garg,
  Behl, and Papailiopoulos}]{shrivastava2025sample}
Vaishnavi Shrivastava, Ahmed~H Awadallah, Vidhisha Balachandran, Shivam Garg,
  Harkirat Behl, and Dimitris Papailiopoulos. 2026.
\newblock Sample more to think less: Group filtered policy optimization for
  concise reasoning.
\newblock In \emph{International Conference on Learning Representations},
  volume 2026, pages 98439--98462.

\bibitem[{Tang et~al.(2025)Tang, Liu, Ye, Lu, Wei, Wang, Lin, Feng, Zhao, Wang
  et~al.}]{tang2025mtvqa}
Jingqun Tang, Qi~Liu, Yongjie Ye, Jinghui Lu, Shu Wei, An-Lan Wang, Chunhui
  Lin, Hao Feng, Zhen Zhao, Yanjie Wang, et~al. 2025.
\newblock Mtvqa: Benchmarking multilingual text-centric visual question
  answering.
\newblock In \emph{Findings of the Association for Computational Linguistics:
  ACL 2025}, pages 7748--7763.

\bibitem[{Team et~al.(2025)Team, Du, Yin, Xing, Qu, Wang, Chen, Zhang, Du, Wei
  et~al.}]{team2025kimi}
Kimi Team, Angang Du, Bohong Yin, Bowei Xing, Bowen Qu, Bowen Wang, Cheng Chen,
  Chenlin Zhang, Chenzhuang Du, Chu Wei, et~al. 2025.
\newblock Kimi-vl technical report.
\newblock \emph{arXiv preprint arXiv:2504.07491}.

\bibitem[{Vivoli et~al.(2024)Vivoli, Bertini, and Karatzas}]{vivoli2024comix}
Emanuele Vivoli, Marco Bertini, and Dimosthenis Karatzas. 2024.
\newblock Comix: A comprehensive benchmark for multi-task comic understanding.
\newblock \emph{Advances in Neural Information Processing Systems},
  37:140828--140846.

\bibitem[{Vivoli et~al.(2025)Vivoli, Llabr{\'e}s, Souibgui, Bertini, Llobet,
  and Karatzas}]{vivoli2025comicspap}
Emanuele Vivoli, Artemis Llabr{\'e}s, Mohamed~Ali Souibgui, Marco Bertini,
  Ernest~Valveny Llobet, and Dimosthenis Karatzas. 2025.
\newblock Comicspap: understanding comic strips by picking the correct panel.
\newblock In \emph{International Conference on Document Analysis and
  Recognition}, pages 337--350. Springer.

\bibitem[{Wang et~al.(2026)Wang, Li, Huang, Wang, Wang, Zhang, zheng, Bai,
  Kang, Feng, Wang, and Zhang}]{wang2025traceable}
Haochen Wang, Xiangtai Li, Zilong Huang, Anran Wang, Jiacong Wang, Tao Zhang,
  Jiani zheng, Sule Bai, Zijian Kang, Jiashi Feng, Zhuochen Wang, and Zhaoxiang
  Zhang. 2026.
\newblock Traceable evidence enhanced visual grounded reasoning: Evaluation and
  method.
\newblock In \emph{International Conference on Learning Representations},
  volume 2026, pages 148769--148794.

\bibitem[{Wang et~al.(2025{\natexlab{a}})Wang, Gao, Gu, Pu, Cui, Wei, Liu,
  Jing, Ye, Shao, Wang, Chen, Zhang, Yang, Wang, Wei, Yin, Li, Cui, Chen, Ding,
  Tian, Wu, Xie, Li, Yang, Duan, Wang, Hou, Hao, Zhang, Li, Zhao, Duan, Deng,
  Fu, He, Wang, He, Shi, He, Xiong, Lv, Wu, Shao, Zhang, Deng, Qi, Ge, Guo,
  Zhang, Zhang, Cao, Lin, Tang, Gao, Huang, Gu, Lyu, Tang, Wang, Lv, Ouyang,
  Wang, Dou, Zhu, Lu, Lin, Dai, Su, Zhou, Chen, Qiao, Wang, and
  Luo}]{wang2025internvl3}
Weiyun Wang, Zhangwei Gao, Lixin Gu, Hengjun Pu, Long Cui, Xingguang Wei,
  Zhaoyang Liu, Linglin Jing, Shenglong Ye, Jie Shao, Zhaokai Wang, Zhe Chen,
  Hongjie Zhang, Ganlin Yang, Haomin Wang, Qi~Wei, Jinhui Yin, Wenhao Li, Erfei
  Cui, and 56 others. 2025{\natexlab{a}}.
\newblock \href {https://arxiv.org/abs/2508.18265} {Internvl3.5: Advancing
  open-source multimodal models in versatility, reasoning, and efficiency}.
\newblock \emph{Preprint}, arXiv:2508.18265.

\bibitem[{Wang et~al.(2025{\natexlab{b}})Wang, Xia, Song, Guan, Dong, Li, Yang,
  Pu, Luo, Wang, Meng, Li, and Sui}]{wang2025beyond}
Xiaochen Wang, Heming Xia, Jialin Song, Longyu Guan, Qingxiu Dong, Rui Li,
  Yixin Yang, Yifan Pu, Weiyao Luo, Yiru Wang, Xiangdi Meng, Wenjie Li, and
  Zhifang Sui. 2025{\natexlab{b}}.
\newblock Beyond single frames: Can {LMM}s comprehend implicit narratives in
  comic strip?
\newblock In \emph{Findings of the Association for Computational Linguistics:
  EMNLP 2025}, pages 6436--6452, Suzhou, China. Association for Computational
  Linguistics.

\bibitem[{Xiao et~al.(2025)Xiao, Huang, Qiu, Tao, Yang, Hong, Wang, and
  Yao}]{xiao2025egoblind}
Junbin Xiao, Nanxin Huang, Hao Qiu, Zhulin Tao, Xun Yang, Richang Hong, Meng
  Wang, and Angela Yao. 2025.
\newblock Egoblind: Towards egocentric visual assistance for the blind.
\newblock In \emph{Advances in Neural Information Processing Systems}, volume
  38, Main Conference. Curran Associates, Inc.

\bibitem[{Yang et~al.(2025{\natexlab{a}})Yang, Wen, Ding, Liu, Chu, Song, Rao,
  Yi, Li, Zang et~al.}]{yang2025kwai}
Biao Yang, Bin Wen, Boyang Ding, Changyi Liu, Chenglong Chu, Chengru Song,
  Chongling Rao, Chuan Yi, Da~Li, Dunju Zang, et~al. 2025{\natexlab{a}}.
\newblock Kwai keye-vl 1.5 technical report.
\newblock \emph{arXiv preprint arXiv:2509.01563}.

\bibitem[{Yang et~al.(2025{\natexlab{b}})Yang, Han, Luo, and
  Hovy}]{yang2025magic}
Shuo Yang, Caren Han, Siwen Luo, and Eduard Hovy. 2025{\natexlab{b}}.
\newblock Magic-vqa: Multimodal and grounded inference with commonsense
  knowledge for visual question answering.
\newblock In \emph{Findings of the Association for Computational Linguistics:
  ACL 2025}, pages 16967--16986.

\bibitem[{Yu et~al.(2025{\natexlab{a}})Yu, Zhang, Zhu, Yuan, Zuo, Yue, Dai,
  Fan, Liu, liu, Liu, Liu, Lin, Lin, Ma, Sheng, Tong, Zhang, Zhang, Zhang,
  Zhang, Zhu, Zhu, Chen, Chen, Wang, Yu, Song, Wei, Zhou, Liu, Ma, Zhang, Yan,
  Wu, and Wang}]{yu2025dapo}
Qiying Yu, Zheng Zhang, Ruofei Zhu, Yufeng Yuan, Xiaochen Zuo, Yu~Yue, Weinan
  Dai, Tiantian Fan, Gaohong Liu, juncai liu, LingJun Liu, Xin Liu, Haibin Lin,
  Zhiqi Lin, Bole Ma, Guangming Sheng, Yuxuan Tong, Chi Zhang, Mofan Zhang, and
  17 others. 2025{\natexlab{a}}.
\newblock {DAPO}: An open-source llm reinforcement learning system at scale.
\newblock In \emph{Advances in Neural Information Processing Systems}, volume
  38, Main Conference, pages 113222--113244. Curran Associates, Inc.

\bibitem[{Yu et~al.(2025{\natexlab{b}})Yu, Wang, Wang, Huang, Ma, He, Cai,
  Chen, Huang, Zhao et~al.}]{yu2025minicpm}
Tianyu Yu, Zefan Wang, Chongyi Wang, Fuwei Huang, Wenshuo Ma, Zhihui He,
  Tianchi Cai, Weize Chen, Yuxiang Huang, Yuanqian Zhao, et~al.
  2025{\natexlab{b}}.
\newblock Minicpm-v 4.5: Cooking efficient mllms via architecture, data, and
  training recipe.
\newblock \emph{arXiv preprint arXiv:2509.18154}.

\bibitem[{Zhang et~al.(2025)Zhang, Wu, Zhang, Zhao, and Bian}]{zhang2025right}
Qingyang Zhang, Haitao Wu, Changqing Zhang, Peilin Zhao, and Yatao Bian. 2025.
\newblock Right question is already half the answer: Fully unsupervised llm
  reasoning incentivization.
\newblock In \emph{Advances in Neural Information Processing Systems}, volume
  38, Main Conference, pages 67345--67372. Curran Associates, Inc.

\bibitem[{Zheng et~al.(2025)Zheng, Liu, Li, Chen, Yu, Gao, Dang, Liu, Men, Yang
  et~al.}]{zheng2025group}
Chujie Zheng, Shixuan Liu, Mingze Li, Xiong-Hui Chen, Bowen Yu, Chang Gao, Kai
  Dang, Yuqiong Liu, Rui Men, An~Yang, et~al. 2025.
\newblock Group sequence policy optimization.
\newblock \emph{arXiv preprint arXiv:2507.18071}.

\bibitem[{Zheng et~al.(2026)Zheng, Yang, Hong, Zhao, Xu, Yang, Shen, and
  XingYu}]{zheng2025deepeyes}
Ziwei Zheng, Minghao Yang, Jack Hong, Chenxiao Zhao, Guohai Xu, Le~Yang, Chao
  Shen, and XingYu. 2026.
\newblock Deepeyes: Incentivizing "thinking with images" via reinforcement
  learning.
\newblock In \emph{International Conference on Learning Representations},
  volume 2026, pages 126775--126798.

\bibitem[{Zhou et~al.(2025)Zhou, Feng, Zhu, Yao, Koyejo, and Han}]{zhoupassive}
Zhanke Zhou, Xiao Feng, Zhaocheng Zhu, Jiangchao Yao, Sanmi Koyejo, and Bo~Han.
  2025.
\newblock From passive to active reasoning: Can large language models ask the
  right questions under incomplete information?
\newblock In \emph{International Conference on Machine Learning}, pages
  78714--78758. PMLR.

\end{thebibliography}

\appendix

\section{Prompting Framework}
\label{sec:append_prompt}

\method{} uses a two-stage prompting framework during policy optimization. 
The first stage samples \activesketch{} rollouts, where the policy produces a compact panel-grounded evidence sketch followed by a final answer. The second stage estimates answer-level preference by asking the policy model to rank distinct candidate answers sampled from the first stage. Gold answers are not provided in either stage.

\subsection{Active Narrative Sketching Prompt}

\Cref{fig:stage1_prompt} shows the prompt used to sample \activesketch{} rollouts from the policy model. Given a manga page with visible panel IDs and a question, the model is instructed to actively trace evidence across panels before answering. Each reasoning step associates one panel ID with one concise clue, making the generated sketch directly parseable into panel-clue pairs. The prompt also discourages page-level summaries and unrelated visual descriptions, so that the intermediate reasoning remains compact and question-directed.

The output is parsed into two fields. The content inside \texttt{<think>} is used to extract the ordered panel trajectory, while the content inside \texttt{<answer>} is treated as the final answer and later used for self-ranking.

\begin{figure*}[t]
    \centering
    \begin{promptnewbox}{\normalsize Active Narrative Sketching Prompt}
    \normalsize
You are answering a \mangavqa{} question by actively tracing evidence across labeled panels.\\
Goal:\\
Read the page like a manga reader who investigates the answer step by step. Instead of selecting all useful panels first, build an evidence chain as the reasoning unfolds.\\
Each step links one panel id to one concise clue, and each new clue should fill a missing piece of evidence.\\
How to trace evidence:\\
1. Start with the panel most directly related to the question.\\
2. From the current panel, extract one concise clue that helps answer the question.\\
3. Decide what evidence is still missing, such as speaker, action, dialogue meaning, emotion, object, location, time, cause, or result.\\
4. Move to the same panel or another panel only if it provides a different clue needed to fill that gap.\\
5. Stop once the collected clues are sufficient to answer the question.
Important:\\
- A panel may be used multiple times if it provides different necessary clues.\\
- Do not describe unrelated details or summarize the whole page.\\
- Use only visible evidence from the manga page.\\
- Keep the final answer brief and do not include the evidence chain in it.\\
Question:\\
\{question\}\\
Output format:\\
<think>\\
pX: clue $\to$ pY: clue $\to$ $\ldots$\\
</think>\\
<answer>brief final answer</answer>
    \end{promptnewbox}
    \caption{Prompt used for Active Narrative Sketching.}
    \label{fig:stage1_prompt}
\end{figure*}

\subsection{Rollout Answer Ranking Prompt}

\Cref{fig:stage2_prompt} shows the prompt used to estimate answer-level preference without gold answers. For each training instance, we first extract the final answers from sampled \activesketch{} rollouts and merge duplicate answers into distinct candidate options. The policy is then asked to rank these options from best to worst according to answer relevance, visual support, completeness, and conciseness. The resulting rankings are aggregated with Borda count to produce the answer preference reward.

This stage produces a relative ordering over candidate answers rather than an absolute correctness label. We sample multiple rankings to reduce ranking noise and obtain dense preference scores for all candidate answers.

\begin{figure*}[t]
    \centering
    \begin{promptnewbox}{\normalsize Rollout Answer Ranking Prompt}
    \normalsize
You are judging candidate answers for a \mangavqa{} question.\\
Rank the options from best to worst. A better answer should:\\
1. Directly answer the question.\\
2. Be fully supported by the manga image.\\
3. Include all necessary information without extra explanation.\\
4. Avoid unsupported details, vague wording, and reasoning text.\\
Important:\\
- Do not reward an answer for being longer.\\
- If two answers are similarly correct, prefer the more concise and direct one.\\
- Penalize extra details that are not needed or not clearly supported by the image.\\
Question: \{question\}\\
Options:\\
\{option\_lines\}\\
Output:\\
Return all option letters exactly once, from best to worst, separated by commas.
    \end{promptnewbox}
    \caption{Prompt used for Rollout Answer Ranking.}
    \label{fig:stage2_prompt}
\end{figure*}

\section{Benchmark \& Models}

\subsection{Manga Understanding Benchmark}

We evaluate model performance on three manga understanding benchmarks: MangaVQA, ChrOMIC, and MangaUB. The detailed information about these benchmarks is as follows:

\begin{itemize}
    \item MangaVQA~\citep{baek2026mangavqa}: A benchmark designed to evaluate multimodal large language models on contextual visual question answering over manga pages. Built on images from Manga109~\citep{aizawa2020building}, it consists of 526 manually curated and verified question–answer pairs that require models to jointly understand visual content, panel layout, character interactions, and embedded textual information. The benchmark covers both single-panel and multi-panel reasoning, and includes questions requiring either exact text extraction or descriptive contextual interpretation.
    \item ChrOMIC~\citep{hou2026chromic}: A benchmark designed to evaluate chronological reasoning in multi-panel comics. Built from public comic datasets, it contains 998 comic pages spanning Western and Japanese styles. The benchmark defines three complementary tasks including panel reordering, description Reordering and multiple-choice question answering. Through a human–AI collaborative annotation pipeline with manual refinement and filtering, ChrOMIC provides a focused testbed for assessing whether LVLMs can move beyond static image understanding toward layout-aware, temporally grounded narrative reasoning in comics.
    \item MangaUB~\citep{ikuta2025mangaub}: A manga understanding benchmark built on Manga109 to evaluate large multimodal models across single- and multi-panel scenarios. It contains 6,585 questions and 18,179 prompts covering both low-level scene recognition and higher-level narrative understanding, including tasks such as location, time, weather, character counting, emotion recognition, panel localization, and next-panel prediction. By combining expert annotations with multiple-choice prompt variants, MangaUB provides a compact testbed for assessing models’ visual, contextual, and sequential understanding of manga.
\end{itemize}

\subsection{General Perception Models}

We evaluated and measured both manga-specialized and general-purpose multimodal large language models (MLLM) on the above benchmarks:

\begin{itemize}
    \item \textbf{MangaLMM}~\citep{baek2026mangavqa}: A manga-specialized VLM introduced together with MangaVQA. It is adapted from Qwen2.5-VL and fine-tuned on MangaOCR annotations and synthetic manga VQA data, making it the most closely related domain-specific baseline for manga visual question answering.
    \item \textbf{Molmo2-8B}~\citep{clark2026molmo2}: An open-weight VLM from Ai2 designed for image, multi-image, and video understanding with grounding capabilities. It uses an efficient packing and message-tree encoding recipe, bidirectional attention over vision tokens, and a token-weighting strategy to improve multimodal grounding and video-language understanding.
    \item \textbf{InternVL3.5-8B}~\citep{wang2025internvl3}: An open-source MLLM from the InternVL3.5 family. Its training pipeline combines multimodal continual pre-training, supervised fine-tuning, and Cascade Reinforcement Learning, with the goal of improving multimodal reasoning, document/OCR understanding, multi-image comprehension, and general visual-language instruction following.
    \item \textbf{MiniCPM-V-4.5}~\citep{yu2025minicpm}: An efficient general-purpose MLLM from the MiniCPM-V series. It is built on Qwen3-8B and SigLIP2-400M, and introduces a unified 3D-Resampler for compact image and video encoding, together with training strategies for document understanding, OCR, and both short- and long-form reasoning.
    \item \textbf{Keye-VL-1.5-8B}~\citep{yang2025kwai}: A general-purpose multimodal model developed by the Kwai Keye Team. It uses Qwen3-8B as the language decoder and a SigLIP-initialized vision encoder, and supports native dynamic resolution, SlowFast video encoding, and 3D RoPE for unified image, video, and text processing.
    \item \textbf{LLaVA-OneVision-1.5}~\citep{an2025llava}: A fully open-source VLM from the LLaVA-OneVision-1.5 family. It is trained with a released end-to-end framework and large-scale curated pretraining and instruction-tuning data, supporting native-resolution image understanding and general multimodal instruction following.
    \item \textbf{Kimi-VL-A3B-Instruct}~\citep{team2025kimi}: An efficient open-source MoE VLM developed by Moonshot AI. It activates only a small subset of language-model parameters during inference while supporting multimodal reasoning, long-context understanding, and agent-oriented capabilities.
    \item \textbf{Qwen2.5-VL-7B-Instruct}~\citep{bai2025qwen25vltechnicalreport}: A general-purpose VLM from the Qwen2.5-VL series. It improves visual recognition, object localization, document parsing, long-video understanding, and visual-agent capabilities, and serves as the backbone for several active visual reasoning baselines in our experiments.
    \item \textbf{Qwen3-VL-8B-Instruct}~\citep{bai2025qwen3}: A general-purpose VLM from the Qwen3-VL series. It supports long interleaved multimodal contexts and introduces architectural updates such as Interleaved-MRoPE, DeepStack, and text-timestamp alignment to improve spatial-temporal modeling, fine-grained vision-language alignment, and video temporal grounding.
\end{itemize}

\subsection{Active Perception Models}

We further evaluated models that go beyond general perception by explicitly selecting or inspecting visual regions during reasoning:

\begin{itemize}
    \item \textbf{TreeVGR}~\citep{wang2025traceable} is initialized from Qwen2.5-VL-7B and trained with a two-stage pipeline consisting of cold-start supervised fine-tuning and reinforcement learning. It introduces traceable visual evidence by supervising both answer correctness and bounding-box localization quality, encouraging the model to ground its reasoning in relevant image regions.
    \item \textbf{DeepEyes}~\citep{zheng2025deepeyes} is built on Qwen2.5-VL-7B and learns active visual reasoning through end-to-end reinforcement learning. During reasoning, it can ground and crop relevant image regions, forming an interleaved multimodal reasoning process that supports fine-grained visual inspection.
    \item \textbf{ViGoRL}~\citep{sarch2025grounded} is also based on Qwen2.5-VL and trains visually grounded reasoning with reinforcement learning. It anchors reasoning steps to explicit image coordinates and uses a multi-turn visual feedback mechanism to dynamically zoom into predicted regions for fine-grained perception.
\end{itemize}

\subsection{Software and Inference Settings}

For \method{}, we implement training with PyTorch 2.8.0, Transformers 4.57.0, Verl 0.7.1, and vLLM 0.11.0. For baseline models, we follow their official repositories and use their recommended software environments, preprocessing pipelines, and default inference configurations unless otherwise specified.

\section{Extended Results}
\label{sec:append_result}

\subsection{Inference Efficiency}

To further quantify inference efficiency, we measure per-example inference latency on MangaVQA using a single NVIDIA A800 GPU with a batch size of 1. As shown in \Cref{tab:latency}, \method{} requires 5.21 seconds per example, lower than all existing active-perception baselines. In particular, it is substantially faster than DeepEyes and ViGoRL, which incur additional overhead from iterative cropping and visual feedback. Although \method{} performs iterative panel selection and evidence construction, its latency remains comparable to that of single-pass models while delivering stronger MangaVQA performance. This favorable efficiency--performance trade-off is consistent with the compact design of ANS, which retains only concise, question-relevant clues during inference.

\begin{table}[!]
\centering
\small
\begin{tabular}{lc}
\toprule
\textbf{Models} & \textbf{Latency (s)} \\
\midrule
\rowcolor{sectiongray}
\multicolumn{2}{c}{\textit{General Perception}} \\
\midrule
Qwen3-VL-8B-Instruct & 8.09 \\
LLaVA-OneVision-1.5  & 3.46 \\
Kimi-VL-A3B-Instruct & 4.16 \\
\midrule
\rowcolor{sectiongray}
\multicolumn{2}{c}{\textit{Active Perception}} \\
\midrule
TreeVGR              & 5.88 \\
DeepEyes             & 30.02 \\
ViGoRL               & 70.49 \\
\rowcolor{oursblue}
\method{}            & \textbf{5.21} \\
\bottomrule
\end{tabular}
\caption{Per-example inference latency on MangaVQA, measured on a single NVIDIA A800 GPU with a batch size of 1. The best result among active-perception methods is shown in bold.}
\label{tab:latency}
\end{table}

\subsection{Comparison with Closed-Source Models}
\label{sec}

To provide broader reference points, we additionally evaluate several strong closed-source models using the same evaluation protocol. As shown in \Cref{tab:closed_source}, \method{} remains competitive despite using an 8B open-source backbone, outperforming several closed-source models on both benchmarks. While the strongest closed-source models achieve higher overall performance, these results demonstrate that \method{} provides a competitive alternative through targeted post-training for manga narrative understanding.

\begin{table}[!]
\centering
\small
\begin{tabular}{lcc}
\toprule
\textbf{Models} & \textbf{MangaVQA} & \textbf{ChrOMIC (MC)} \\
\midrule
Claude Sonnet 4.5 & 5.84 & 0.4762 \\
Claude Sonnet 5   & 6.48 & 0.6017 \\
GPT-5.2           & 6.59 & 0.5281 \\
GPT-5.4 mini      & 6.25 & 0.4459 \\
GPT-5.4           & 7.98 & 0.5714 \\
Gemini 2.5 Flash  & 7.26 & 0.5974 \\
Gemini 3.5 Flash  & \textbf{8.33} & \textbf{0.7229} \\
\midrule
\rowcolor{oursblue}
\method{} & 6.90 & 0.5801 \\
\bottomrule
\end{tabular}
\caption{Comparison with strong closed-source models under the same evaluation protocol. MangaVQA open-ended responses are scored from 1 to 10 using the LLM judge, while ChrOMIC reports accuracy in its original multiple-choice setting.}
\label{tab:closed_source}
\end{table}

\subsection{Evidence Behavior of \activesketch{}}

\begin{figure}[!h]
    \centering
    \includegraphics[width=0.9\columnwidth]{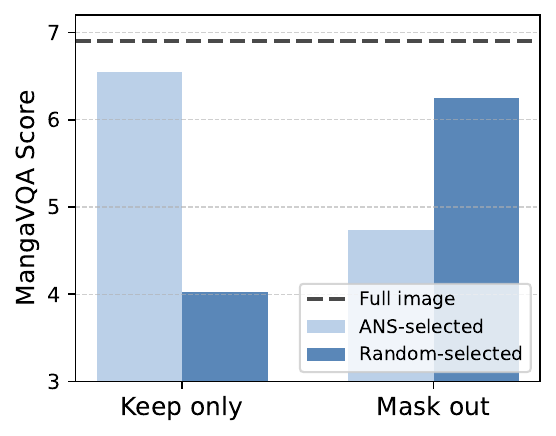}
    \caption{
        Counterfactual panel perturbation analysis on MangaVQA. We compare \activesketch{}-selected and randomly selected panels under keep-only and mask-out settings. The dashed line denotes the full-image score.
    }
    \label{fig:mask}
\end{figure}

The process analysis in the main text compares \method{} with other active perception methods. Here, we further focus on the behavior of \activesketch{} itself and examine whether its generated sketches capture useful evidence and organize it in an effective way.

\paragraph{Panel-level evidence relevance.}
We first test whether the panels selected by \activesketch{} are actually useful for answering. As shown in~\Cref{fig:mask}, keeping only \activesketch{}-selected panels yields much better performance than keeping the same number of randomly selected panels. Conversely, masking \activesketch{}-selected panels leads to a larger degradation than masking random panels. These two perturbations show that \activesketch{}-selected panels are strongly associated with answer-relevant evidence. Notably, the generated evidence path contains only 2.4 panels on average out of 11.3 panels per page, indicating that \activesketch{} preserves substantial answer-related information with a compact panel subset.

\begin{figure}[!h]
    \centering
    \includegraphics[width=0.9\columnwidth]{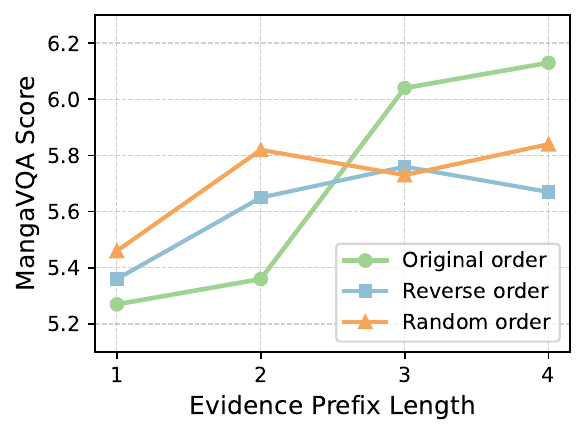}
    \caption{
        Evidence sketch accumulation analysis. We reorder length-4 evidence sketch generated by \activesketch{} into original, reverse, and random orders, progressively reveal their prefixes, and report the MangaVQA score at each prefix length.
    }
    \label{fig:accum}
\end{figure}

\paragraph{Evidence sketch accumulation.}
Beyond selecting relevant panels, we examine whether the organization of \activesketch{} affects answer generation. For model outputs on MangaVQA whose sketch contains four steps, we keep the steps in their original order or reorder them into reverse and random orders. For each order, we truncate the sketch after the first $k$ steps ($k=1,\ldots,4$), append the resulting prefix to the original prompt, and ask the model to generate the final answer.

As shown in~\Cref{fig:accum}, the original \activesketch{} produces a steadily improving curve as more steps are revealed, while the reverse and random variants saturate or fluctuate. Since all variants contain the same panel-clue steps, the difference cannot be explained by evidence selection alone. Instead, it suggests that \activesketch{} is useful not only because it collects relevant clues, but also because it arranges them into a question-directed evidence flow that better supports manga narrative understanding.

\end{document}